\documentclass[10pt,twocolumn,letterpaper]{article}

\usepackage{cvpr}              
\definecolor{cvprblue}{rgb}{0.21,0.49,0.74}
\usepackage[pagebackref,breaklinks,colorlinks,allcolors=cvprblue]{hyperref}
\usepackage{pifont}
\usepackage{makecell}
\usepackage{amssymb}
\usepackage{multirow}
\usepackage[table]{xcolor}

\def\paperID{44181} 
\def\confName{CVPR}
\def\confYear{2026}
\definecolor{Blue}{RGB}{0, 119, 182}

\title{\textcolor{Blue}{Eevee}: Towards Clos\textcolor{Blue}{e}-up High-r\textcolor{Blue}{e}solution \textcolor{Blue}{V}id\textcolor{Blue}{e}o-bas\textcolor{Blue}{e}d Virtual Try-on}

\author{
    Jianhao Zeng\textsuperscript{1}\footnotemark[1] \quad
    Yancheng Bai\textsuperscript{1}\footnotemark[1] \footnotemark[2] \quad
    Ruidong Chen\textsuperscript{1,2} \quad
    Xuanpu Zhang\textsuperscript{2} \quad
    Lei Sun\textsuperscript{1} \\
    Dongyang Jin\textsuperscript{1} \quad
    Ryan Xu\textsuperscript{1} \quad
    Nannan Zhang\textsuperscript{3}\footnotemark[3]  \quad
    Dan Song\textsuperscript{2} \quad
    Xiangxiang Chu\textsuperscript{1} \\
    \textsuperscript{1}Amap, Alibaba Group \quad
    \textsuperscript{2}Tianjin University \\
    \textsuperscript{3}Shenzhen Institutes of Advanced Technology, Chinese Academy of Sciences  \\
}

\begin{document}
\twocolumn[{
\renewcommand\twocolumn[1][]{#1}
\maketitle
    \includegraphics[width=\linewidth]{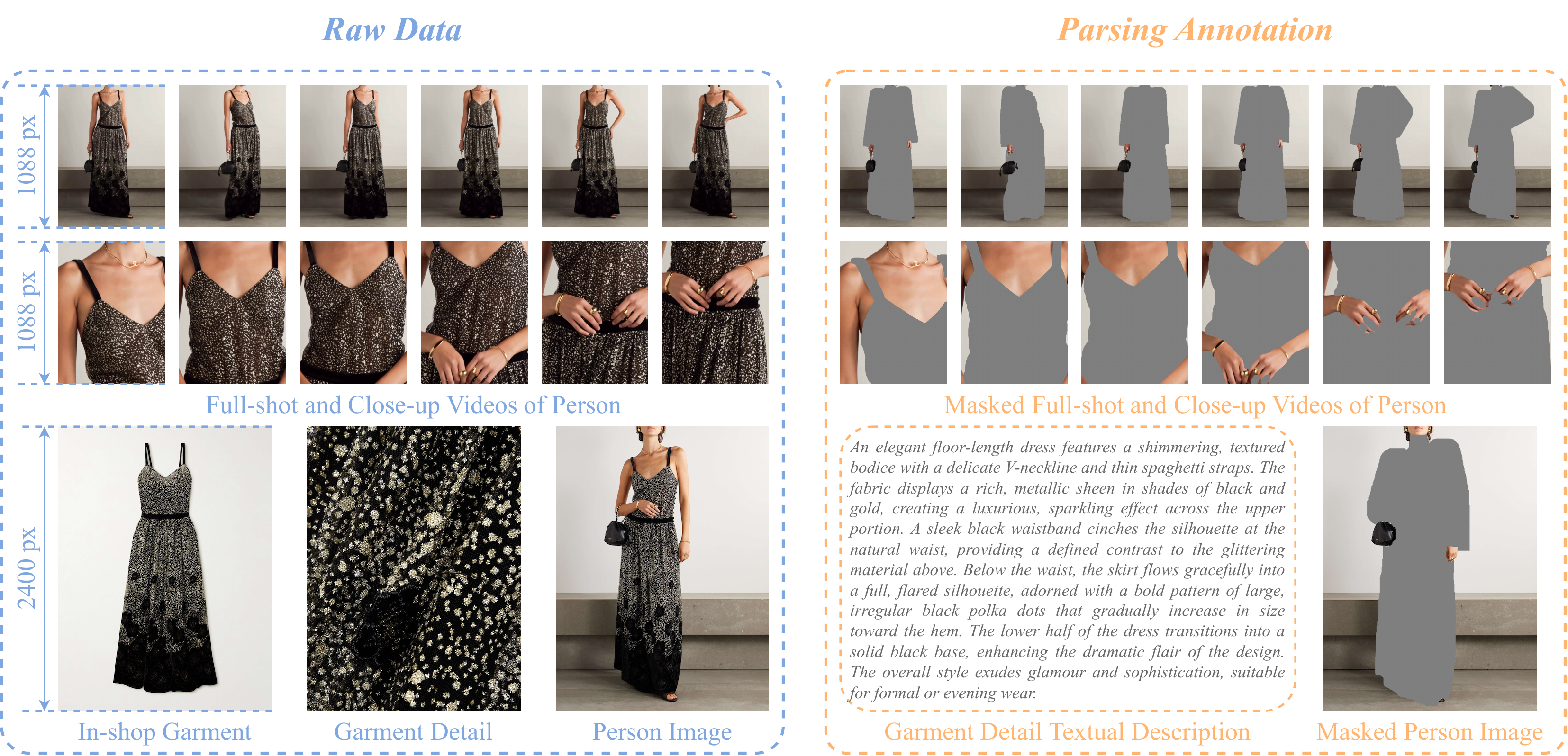}
    \captionof{figure}{
    An illustrative sample from our proposed virtual try-on dataset. Compared to existing datasets, ours provides a richer collection of data, including both full-shot and close-up videos of the person, detailed garment images, and rich textual descriptions. Furthermore, all images and videos are provided at a higher resolution. More examples can be found in the Appendix.
    }
    \label{Fig:Teaser}
    \vspace{3mm}
}]

\begin{abstract}
\renewcommand{\thefootnote}{\fnsymbol{footnote}}
\footnotetext[1]{Equal contribution.}
\footnotetext[2]{Project Lead.}
\footnotetext[3]{Corresponding author: Nannan Zhang (nn.zhang2@siat.ac.cn). }
\footnotetext[4]{The research is supported in part by the National Natural Science Foundation of China (62571369). The code is available at \url{https://github.com/AMAP-ML/Eevee}.}

Video virtual try-on technology provides a cost-effective solution for creating marketing videos in fashion e-commerce.
However, its practical adoption is hindered by two critical limitations. First, the reliance on a single garment image as input in current virtual try-on datasets limits the accurate capture of realistic texture details. Second, most existing methods focus solely on generating full-shot virtual try-on videos, neglecting the business's demand for videos that also provide detailed close-ups.
To address these challenges, we introduce a high-resolution dataset for video-based virtual try-on. This dataset offers two key features. First, it provides more detailed information on the garments, which includes high-fidelity images with detailed close-ups and textual descriptions; Second, it uniquely includes full-shot and close-up try-on videos of real human models.
Furthermore, accurately assessing consistency becomes significantly more critical for the close-up videos, which demand high-fidelity preservation of garment details.
To facilitate such fine-grained evaluation, we propose a new garment consistency metric VGID (Video Garment Inception Distance) that quantifies the preservation of both texture and structure.
Our experiments validate these contributions. We demonstrate that by utilizing the detailed images from our dataset, existing video generation models can extract and incorporate texture features, significantly enhancing the realism and detail fidelity of virtual try-on results.
Furthermore, we conduct a comprehensive benchmark of recent models. The benchmark effectively identifies the texture and structural preservation problems among current methods.

\end{abstract}
\section{Introduction}

Driven by the rapid growth of e-commerce, the demand for apparel marketing videos is surging, making virtual try-on technology~\cite{MEF, CATDM, Fashion, Robust} a crucial content generation tool. A foundational task in this domain is video-based virtual try-on~\cite{CATV2TON, Clothformer, MagicTryon, VIVID, RealVVT}, where the goal is to synthesize a video from a target garment image and a person's source video. In the resulting video, the individual is realistically dressed in the target garment while maintaining their original movements and identity.

However, due to dataset limitations, existing video-based virtual try-on models face several common technical hurdles in practical applications.
First, existing datasets typically only provide a single full-shot video as the ground truth. This results in videos that are more like a general, low-detail preview, failing to capture how the garment truly behaves and failing to meet the business demand for detailed close-up videos.
Consequently, the practical value of this technology is significantly limited.
Moreover, constrained by the low resolution of training datasets like ViViD~\cite{VIVID} ($832\times624$) and VVT~\cite{FWGAN} ($256\times192$), existing models~\cite{DreamVVT, DPIDM, MagicTryon, VitonDIT} produce videos with insufficient clarity. This coarse visual output fails to provide users with adequate detail to assess the garment's true material and fit.
Additionally, the majority of these models rely on a single, flat image of the garment as input, making it challenging to generate lifelike fabric folds and intricate details. In practice, however, vendors can readily supply multi-angle detail shots of their garments, yet this valuable information remains largely underutilized.

\begin{figure}[t]
  \centering
   \includegraphics[width=\linewidth]{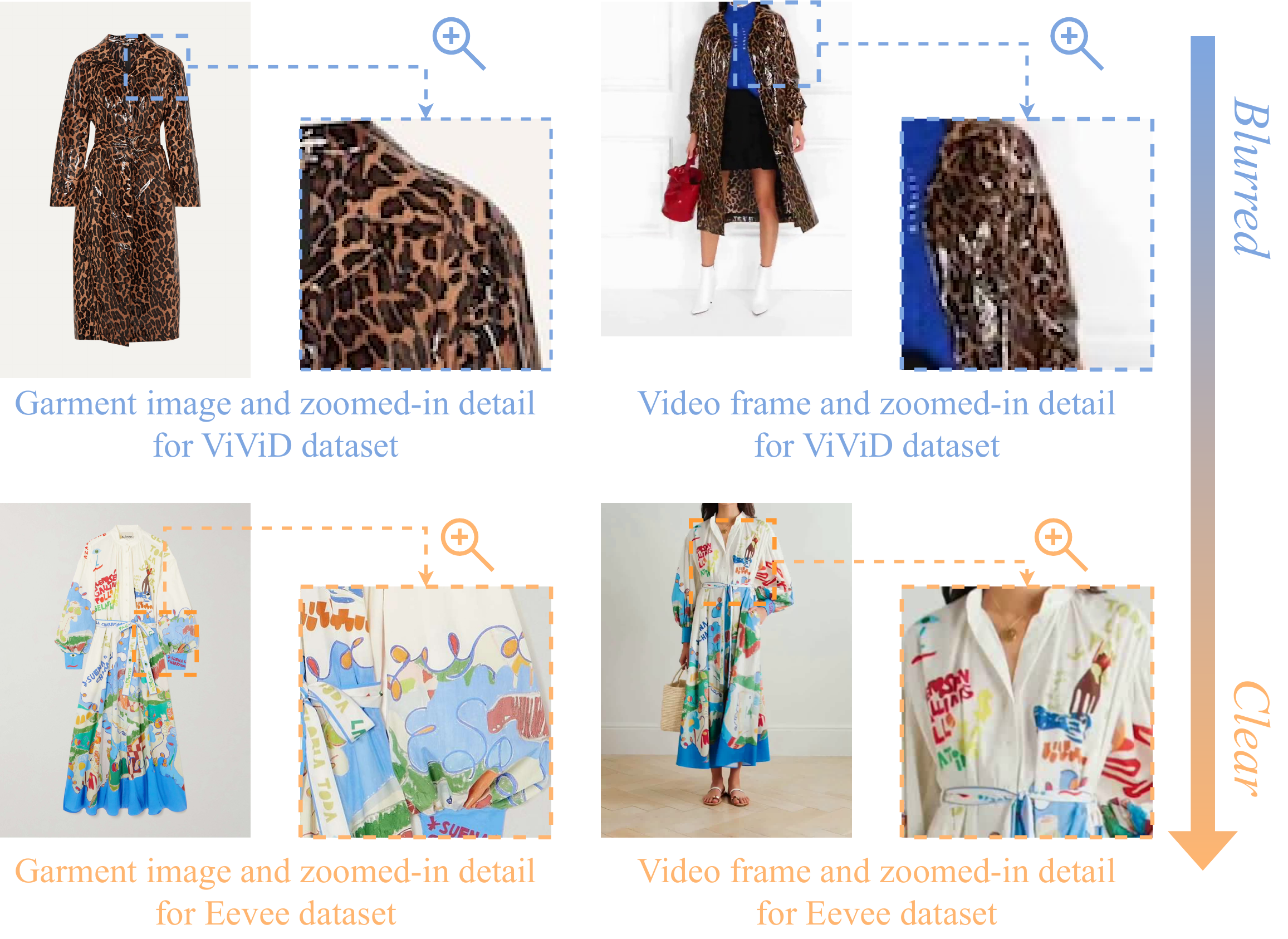}

   \caption{A comparison of garment image and video frame, along with their zoomed-in details, from the ViViD dataset shown on top and the Eevee dataset shown on bottom.}
   \label{Fig:DatasetDetail}
\end{figure}

To address the aforementioned issues, we introduce Eevee, a high-resolution dataset designed to support a wide range of virtual try-on tasks.
Compared to previous datasets, Eevee's main distinctions are as follows:
First, in addition to providing a full-shot video of a person, it also offers a close-up video.
Second, the dataset includes detailed images and a detailed textual description of the garment to assist the model in generating fine details and textures.
Finally, Eevee features the highest resolution currently available for this task.
The practical benefit of this is evident in our comparative analysis: as illustrated in ~\cref{Fig:DatasetDetail}, a magnified, side-by-side comparison with the ViViD dataset shows that Eevee provides substantially clearer details and superior texture fidelity, overcoming the blur common in lower-resolution benchmarks.
\cref{Fig:Teaser} illustrates a sample from our dataset. The left panel displays the raw collected data; the right panel shows its corresponding parsing annotation.
To enhance usability, we applied further data processing beyond this annotation step.

Close-up virtual try-on videos, where the garment occupies the majority of the frame, necessitate a more precise evaluation of garment fidelity. Existing metrics such as VFID~\cite{VFID} and VBench~\cite{Vbench} primarily assess overall video quality and realism, but they are insufficient for gauging the consistency of the generated apparel with the input garment image. To bridge this gap, we propose a new metric, VGID (Video Garment Inception Distance), specifically designed to measure garment consistency. This metric operates by calculating the cosine similarity between deep features extracted from the generated garment region in the video and those from the input garment image.

Leveraging our Eevee dataset, we conduct a comprehensive benchmark of recent state-of-the-art models~\cite{VIVID, VACE, MagicTryon} using our proposed VGID metric alongside established metrics such as VFID, SSIM~\cite{SSIM}, LPIPS~\cite{LPIPS}, and VBench.
This evaluation reveals significant disparities in texture and structural preservation among current methods, thereby establishing a strong baseline for future research in high-fidelity virtual try-on.

In summary, we introduce Eevee, a high-resolution virtual try-on dataset featuring close-up videos and detailed garment data for fine texture generation. To evaluate close-up garment consistency where metrics like VFID or VBench fall short, we propose a novel metric, VGID. Furthermore, we establish a strong baseline for future research by comprehensively benchmarking state-of-the-art models using our proposed dataset and metric.

    

\begin{table*}[t]
    \footnotesize
    \centering
    \setlength{\tabcolsep}{6.4pt}
    \begin{tabular}{l ccc cc cccc}
    \toprule
            \multirow{2}{*}[-3px]{\textbf{Dataset}}  & \multirow{2}{*}[-3px]{\makecell{\textbf{Garment} \\ \textbf{Category}}} & \multirow{2}{*}[-3px]{\makecell{\textbf{Detailed} \\ \textbf{Image}}} & \multirow{2}{*}[-3px]{\makecell{\textbf{Close-up} \\ \textbf{Video}}} & \multicolumn{2}{c}{\textbf{Resolution}} & \multicolumn{4}{c}{\textbf{Total number}} \\
        \cmidrule(lr){5-6} 
        \cmidrule(lr){7-10} 
        & & & & \textbf{Image} & \textbf{Video} &  \textbf{Train pairs} & \textbf{Test pairs} & \textbf{Images} & \textbf{Frames} \\
        \midrule
        VITON~\cite{VITON}             & Upper-body & \ding{55} & \ding{55} & $256\times192$  & -- & 14,221 & 2,032  & 32,506 & -- \\
        MPV~\cite{MPV}                 & Upper-body & \ding{55} & \ding{55} & $256\times192$  & -- & 52,336 & 10,544 & 49,211 & -- \\
        VITON-HD~\cite{VITONHD}        & Upper-body & \ding{55} & \ding{55} & $1024\times768$ & -- & 11,647 & 2,032  & 27,358 & -- \\
        DressCode~\cite{Dresscode}     & Multi      & \ding{55} & \ding{55} & $1024\times768$ & -- & 48,392 & 5,400  & 161,376 & -- \\
        DeepFashion~\cite{Deepfashion} & \ding{55}  & \ding{55} & \ding{55} & $1101\times750$ & -- & \multicolumn{2}{c}{52,712} & 52,712 & -- \\
        \midrule
        VVT~\cite{FWGAN}               & Upper-body & \ding{55} & \ding{55} & $256\times192$  & $256\times192$  & 661    & 130    & 1,582 & 205,675   \\
        TikTokDress~\cite{Swifttry}    & Upper-body & \ding{55} & \ding{55} & $720\times540$  & $720\times540$  & 693    & 124    & 817   & 272,548   \\
        ViViD\cite{VIVID}              & Multi      & \ding{55} & \ding{55} & $1380\times920$ & $832\times624$  & 7,759  & 1,941  & 9,700 & 1,213,694 \\
        \rowcolor[rgb]{0.9,0.95,1}
        Eevee (Ours)                   & Multi      & \ding{51} & \ding{51} & $2400\times1800$& $1088\times816$ & 8,364 & 1,000 & 28,092 & 1,905,973  \\
    
    \bottomrule
    \end{tabular}
    \caption{Comparison between our dataset and the most widely used datasets for virtual try-on and other related tasks.}
    \label{Table:Dataset}
\end{table*}

\section{Related Work}

\subsection{Video-based Virtual Try-on}
With the rapid development of Diffusion models~\cite{DDPM, DDIM, Flow, FluxText, Pixel, Peng, Anyscene, LayerBind, GRAG, RL3DEdit, Jasmine, FE2E} and Transformers~\cite{Transformer, Scalar, UniCTokens, Mc-llava, Analyst, Visionllama, Perceive, Usp, Draw, Twins}, video generation~\cite{Cogvideox, opensora} and video editing~\cite{Propainter, VACE, LTX} models have achieved significant success. This progress is now impacting virtual try-on technology, which aims to synthesize a target garment onto a human model to generate try-on results. While significant progress has been made in image-based virtual try-on~\cite{StableVTON, CATDM, Fashion, LADI, DCI,IDM, GPVTON}, research on the more complex video-based virtual try-on is still in its early stages. In this emerging field, researchers have begun their initial explorations. For example, RealVVT~\cite{RealVVT} employs a dual U-Net~\cite{Doubleu-net} architecture to generate precise virtual try-on videos, enhancing both spatial and temporal consistency. CatV2TON~\cite{CATV2TON} is a Diffusion Transformer (DiT)~\cite{Transformer} framework for both image and video virtual try-on, which implements a DiT~\cite{DiT} architecture by temporally concatenating garment and person inputs. Similarly, MagicTryon~\cite{MagicTryon} builds upon the DiT backbone from Wan2.1~\cite{Wan} and fully leverages its self-attention mechanism to achieve unified modeling of spatiotemporal coherence.
Due to dataset limitations, these methods often only generate full-shot, low-resolution virtual try-on videos that also lack detailed garment textures, which restricts their practical application.

\subsection{Virtual Try-on Datasets}
The availability of high-quality datasets is indispensable for the development of virtual try-on.
Early developments in this field were primarily centered on static, image-based collections. Seminal datasets like VITON~\cite{VITON} first established the task by providing paired images of a person and a target in-shop garment. This was later superseded by VITON-HD~\cite{VITONHD}, which significantly increased the image resolution to $1024\times768$, enabling the generation of more realistic and detailed results. Following this, DressCode~\cite{Dresscode} further expanded the scale and complexity, offering a large-scale repository with more diverse poses and challenging garment types.
While effective for static try-on, these image-based datasets lack the temporal information necessary for video-based applications. Efforts to bridge this gap have led to the creation of video datasets, most notably VVT~\cite{FWGAN} and ViViD~\cite{VIVID}.
However, the VVT dataset is of limited utility due to its low video resolution and small scale. Although ViViD represents a significant advancement in terms of resolution and data volume, it possesses critical limitations. First, it provides only a single, full-shot video, failing to meet the commercial demand for detailed close-up videos. Furthermore, it only provides a single, flat-lay image for each garment, which is insufficient for capturing the intricate textures, material properties, and deformation details required for high-fidelity virtual try-on.

\section{Eevee}

\subsection{Dataset}
Existing public video-based virtual try-on datasets suffer from three limitations.
First, existing public datasets for video-based virtual try-on typically only contain video from a single global perspective. However, practical applications often require more diverse perspectives, such as close-ups on specific areas, which these datasets lack.
Additionally, these datasets typically represent a garment with only a single, flat in-shop image. This overlooks a key opportunity in practical applications: merchants can readily provide detailed close-up shots, and leveraging such information could enhance the fine-grained details of the generated results.
Finally, the majority of these public datasets suffer from low resolution, which places a hard ceiling on the visual quality of the virtual try-on, often resulting in blurry outputs where essential garment details are lost.

\begin{figure*}[t]
  \centering
   \includegraphics[width=\linewidth]{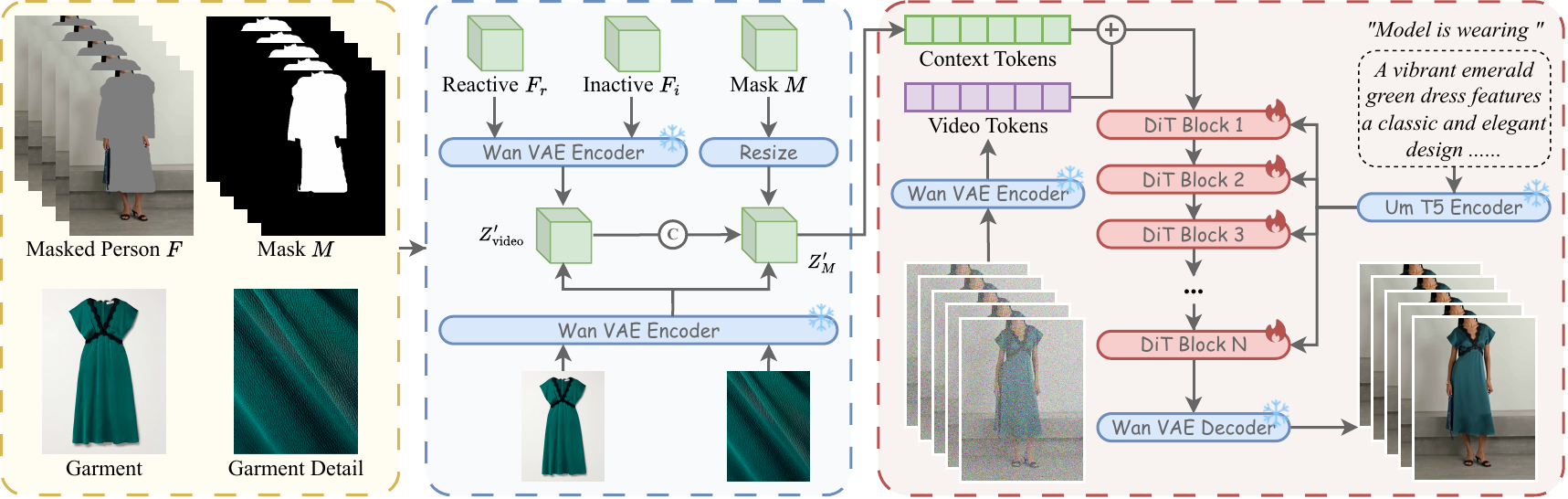}
   \caption{The overall fine-tuning pipeline of the VACE on our dataset. Compared to other video-based virtual try-on models, our VACE fine-tuning process includes an additional step where detailed garment images are fed into the network through the Wan VAE Encoder. }
   \label{Fig:Model}
\end{figure*}

To facilitate the development of video-based virtual try-on, we introduce Eevee. It is a high-resolution dataset designed to overcome prior limitations. It provides both full-shot videos and corresponding close-up videos, addressing the need for detailed dynamic views. Additionally, it moves beyond the single in-shop photo by providing detailed close-up images for each garment. This rich data is designed to support a wide range of virtual try-on tasks.
The Eevee dataset features 9,364 high-resolution, video-based virtual try-on pairs, establishing it as the highest-resolution collection available with images up to $2400 \times 1800$ and videos up to $1088 \times 816$. The dataset is categorized into three main garment types, split into training and testing sets: upper-body (4492 training / 500 test pairs), lower-body (2308 training / 250 test pairs), and dresses (1564 training / 250 test pairs). The total number of images in our dataset, excluding parsing annotations, is 28,092, and the total number of video frames is 1,905,973.

We first collected a variety of images and related video data from e-commerce websites.
The corresponding raw data encompasses in-shop garment images, close-up details of the garment, images of the person, as well as full-shot and close-up videos of the person.
To make the dataset easier to use, we have added a series of annotations.
(1) We utilize the large language model Qwen-VL-Max~\cite{Qwen-VL} to generate detailed textual descriptions and classify garments from images. 
(2) For full-shot virtual try-on videos and person images, we use OpenPose~\cite{Openpose} to obtain human parsing results and then generate the mask. For close-up virtual try-on videos, where obtaining human parsing results is often difficult, we instead use Grounded SAM-2~\cite{GroundedSAM, GroundingDino, SAM2} to perform semantic segmentation on the garment region to extract the mask. We then apply a 31-pixel padding around the mask to mitigate boundary inaccuracies and prevent the leakage of the original garment’s shape information. We also conducte manual random sampling to identify and remove severe segmentation failures.
(3) To adapt to other virtual try-on models, we use AniLines~\cite{AniLines} to extract garment contour maps and Detectron2~\cite{Detectron2} to obtain DensePose UV coordinates for the human body. 
\cref{Fig:Teaser} illustrates a sample from our dataset.

\cref{Table:Dataset} summarizes the key characteristics of our dataset in comparison with existing datasets for virtual try-on and fashion-related tasks. Notably, the Eevee dataset is the first to provide both detailed images of garments and close-up virtual try-on videos. Furthermore, it boasts the highest image and video resolutions currently available.

\subsection{Method}

To validate our dataset, especially the garment detail images, we fine-tune the pre-trained VACE~\cite{VACE} model. VACE, a diffusion transformer-based framework for video generation and editing built on Wan2.1~\cite{Wan}, is ideal for video-based virtual try-on. This task functions as a masked video-to-video editing process.
We chose VACE as our base model for two key reasons. First, it requires only video masks, avoiding the reliance on extra parsing data (like DensePose~\cite{Densepose}) seen in other methods such as MagicTryon~\cite{MagicTryon}. This minimizes interference from annotation inaccuracies. Second, VACE is highly extensible, making it a suitable platform for future work involving more complex multimodal inputs.

The overall fine-tuning pipeline is shown in ~\cref{Fig:Model}. Given an input masked person video $F$ and a mask video $M$, VACE first decouples $F$ using the provided segmentation mask. This decoupling, based on $M$, yields two distinct frame sequences identical in shape:
\begin{equation}
    \begin{split}
    F_r &= F \odot M  \\
    F_i &= F \odot (1-M),
    \end{split}
    \label{eq:1}
\end{equation}
where $F_r$, called reactive frames, contains all the pixels to be changed (such as the gray pixels), while all the pixels to be kept are stored in $F_i$, the inactive frames.

After decoupling, we process these conditional inputs. As shown in the diagram, $F_r$ and $F_i$ are processed independently by the Wan VAE Encoder to produce latent representations, which we denote $Z_r$ and $Z_i$ respectively. The mask $M$ is processed by first being rearranged into patches to create a 64-channel representation, which is then downsampled via interpolation to match the latent dimensions, yielding the mask latent $Z_M$. These processed representations are then used to form the Context Tokens.
Specifically, the inactive latents $Z_i$ and reactive latents $Z_r$ are first concatenated along the channel dimension to create a unified video latent representation $Z_{\text{video}}$.
Next, the garment and garment detail images (treated as $f$ individual reference frames) are encoded by the VAE to produce a set of reference latents $Z_{\text{ref}, j}$. To match the channel structure of $Z_{\text{video}}$, reference latent $Z_{\text{ref},j}$ is concatenated with a zero tensor of the same shape along the channel dimension. These prepared reference latents are then prepended to the main video latent $Z_{\text{video}}$ along the temporal dimension.
Concurrently, the interpolated mask latent $Z_M$ is prepended with a block of zero-latents along the temporal dimension to create $Z'_M$. Finally, $Z'_{\text{video}}$ is concatenated with $Z'_M$ along the channel dimension. This resulting tensor forms the Context Tokens, which are then combined with the Video Tokens and fed into the DiT model.

\subsection{Metric}

Unlike video-based virtual try-on which typically focuses on full-shot video generation, close-up video-based virtual try-on presents a distinct set of challenges. In these zoomed-in perspectives, the garment itself occupies a significantly larger portion of the frame. This shift means that the accurate reconstruction of intricate garment textures and patterns has become the most critical measure of quality. However, existing metrics are ill-equipped for this specific task. For instance, VFID~\cite{VFID} evaluates the overall video quality. Furthermore, while VBench's Subject Consistency~\cite{Vbench} is useful, it primarily measures temporal stability, ensuring that the garment remains consistent within the video. However, it does not assess fidelity to the input garment image. To address this limitation, we propose VGID (Video Garment Inception Distance), a new metric designed to precisely quantify the fidelity between the generated garment texture and the source garment image.

VGID's core idea is to measure the consistency between the generated video's garment and the input garment image. To do this while avoiding interference from the background, a naive approach would be to use segmentation masks. However, to obviate the critical dependency on these unreliable or unavailable masks, VGID leverages the inherent properties of vision transformers. We leverage the DINO-V2~\cite{Dinov2} model, which is renowned for its powerful emergent segmentation capabilities developed during self-supervised training; the model learns to segment objects without explicit supervision. We harness this property by extracting and processing the [CLS]-to-patch self-attention maps from the final Transformer block. This map is then upsampled and normalized to create a soft mask $M$, which assigns high values (near 1.0) to the salient garment and low values (near 0.0) to the background. However, we observe that for categories such as `lower body' and `upper body', the soft masks generated by DINO-V2 frequently fail to isolate the target item, often incorporating parts of adjacent garment pieces. This cross-contamination makes the mask unreliable for evaluating multi-garment outfits. Given that dresses typically function as a single, holistic garment and are less susceptible to this ambiguity, we report this consistency metric exclusively for the `dresses' category to ensure a more accurate and meaningful evaluation.

VGID provides a score, computed as the cosine similarity between the masked, spatially-averaged features of the garment image $I_s$ and a generated frame $I_v$. 
To calculate this, we first extract feature maps $F_s = \Phi(I_s)$ and $F_v = \Phi(I_v)$ using a feature extractor $\Phi$. The corresponding soft masks, $\hat{M}_s$ and $\hat{M}_v$, are resized to match the spatial dimensions of the feature maps and then applied via element-wise multiplication ($F' = F \odot \hat{M}$). This step effectively zeros out background features. The resulting masked features are subjected to global average-pooling (GAP), and their consistency is quantified using cosine similarity, as defined by the equation:
\begin{equation}
    \text{VGID}(I_s, I_v) = \frac{\text{GAP}(F'_s) \cdot \text{GAP}(F'_v)}{\|\text{GAP}(F'_s)\| \|\text{GAP}(F'_v)\|}.
\end{equation}
This calculation forms our semantic fidelity metric, denoted as $\text{VGID}$. It leverages deep, semantically-aware features from DINO-V2 to measure the integrity of high-level patterns—for example, ensuring that a plaid pattern remains recognizably plaid. For a full video $V$ with $T$ frames, the final score is computed as the average across all frames, providing a robust assessment of semantic consistency throughout the video.

\begin{table*}[t]
    \footnotesize
    \centering
    \setlength{\tabcolsep}{7pt}
    
    \begin{tabular}{lll ccc ccccc}
    \toprule
        \multirow{2}{*}[-3px]{\textbf{Resolution}} & \multirow{2}{*}[-3px]{\textbf{Type}} & \multirow{2}{*}[-3px]{\textbf{Method}} & \multicolumn{3}{c}{\textbf{Unpaired}} & \multicolumn{5}{c}{\textbf{Paired}} \\
        \cmidrule(lr){4-6} 
        \cmidrule(lr){7-11} 
         & & & 
        \textbf{$\text{VFID}_\text{R}^\text{u}\downarrow$ } & 
        \textbf{$\text{VFID}_\text{I}^\text{u}\downarrow$ } &
        \textbf{$\text{VGID}^\text{u}\uparrow$ } &
        \textbf{$\text{VFID}_\text{R}^\text{p}\downarrow$ } & 
        \textbf{$\text{VFID}_\text{I}^\text{p}\downarrow$ } & 
        \textbf{$\text{VGID}^\text{p}\uparrow$ } &
        \textbf{SSIM$\uparrow$} &
        \textbf{LPIPS$\downarrow$} \\
    \midrule
        \multirow{4}{*}{$1088\times816$} & \multirow{4}{*}{Full-shot} & ViViD & 0.565 & 12.859 & 0.514 & 0.487 & 10.306 & 0.521 & 0.874 & 0.119 \\
        & & MagicTryon & \textbf{0.187} & \textbf{9.783} & 0.512 & 0.137 & \textbf{6.168} & 0.524 & 0.906 & 0.092 \\
        & & VACE & 0.454 & 11.260 & 0.522 & 0.282 & 7.290 & 0.525 & 0.899 & 0.083 \\
        & & \cellcolor[rgb]{0.9,0.95,1}Ours & 
        \cellcolor[rgb]{0.9,0.95,1}0.220 &
        \cellcolor[rgb]{0.9,0.95,1}10.065 & 
        \cellcolor[rgb]{0.9,0.95,1}\textbf{0.524} & 
        \cellcolor[rgb]{0.9,0.95,1}\textbf{0.076} & 
        \cellcolor[rgb]{0.9,0.95,1}6.555 & 
        \cellcolor[rgb]{0.9,0.95,1}\textbf{0.534} & 
        \cellcolor[rgb]{0.9,0.95,1}\textbf{0.910} & 
        \cellcolor[rgb]{0.9,0.95,1}\textbf{0.078} \\
    \midrule
        \multirow{4}{*}{$832\times624$} & \multirow{4}{*}{Full-shot} & ViViD & 0.389 & 12.194 & 0.506 & 0.285 & 9.211 & 0.508 & 0.871 & 0.111 \\
        & & MagicTryon & \textbf{0.161} & 9.865 & 0.520 & 0.106 & \textbf{6.167} & 0.514 & \textbf{0.908} & 0.074 \\
        & & VACE & 0.587 & 11.647 & 0.524 & 0.343 & 7.624 & 0.528 & 0.888 & 0.090 \\
        & & \cellcolor[rgb]{0.9,0.95,1}Ours & 
        \cellcolor[rgb]{0.9,0.95,1}0.219 & 
        \cellcolor[rgb]{0.9,0.95,1}\textbf{9.777} & 
        \cellcolor[rgb]{0.9,0.95,1}\textbf{0.528} &
        \cellcolor[rgb]{0.9,0.95,1}\textbf{0.076} & 
        \cellcolor[rgb]{0.9,0.95,1}6.520 & 
        \cellcolor[rgb]{0.9,0.95,1}\textbf{0.527} & 
        \cellcolor[rgb]{0.9,0.95,1}0.905 & 
        \cellcolor[rgb]{0.9,0.95,1}\textbf{0.071} \\
    \midrule
        \multirow{4}{*}{$1088\times816$} & \multirow{4}{*}{Close-up} & ViViD & 1.253 & 12.665 & 0.533 & 0.976 & 10.900 & 0.509 & 0.785 & 0.185 \\
        & & MagicTryon & 0.752 & \textbf{11.285} & 0.538 & 0.607 & 8.714 & 0.519 & \textbf{0.807} & 0.163 \\
        & & VACE & 0.913 & 12.844 & 0.533 & 0.429 & 9.331 & 0.515 & 0.778 & 0.174  \\
        & & \cellcolor[rgb]{0.9,0.95,1}Ours &   
        \cellcolor[rgb]{0.9,0.95,1}\textbf{0.535} & 
        \cellcolor[rgb]{0.9,0.95,1}11.621 & 
        \cellcolor[rgb]{0.9,0.95,1}\textbf{0.540} & 
        \cellcolor[rgb]{0.9,0.95,1}\textbf{0.274} & 
        \cellcolor[rgb]{0.9,0.95,1}\textbf{8.429} & 
        \cellcolor[rgb]{0.9,0.95,1}\textbf{0.524} & 
        \cellcolor[rgb]{0.9,0.95,1}0.803 & 
        \cellcolor[rgb]{0.9,0.95,1}\textbf{0.154} \\
    \midrule
        \multirow{4}{*}{$832\times624$} & \multirow{4}{*}{Close-up} & ViViD & 0.936 & 12.198 & 0.533 & 0.663 & 10.116 & 0.509  & 0.775 & 0.176  \\
        & & MagicTryon & 0.595 & \textbf{11.262} & 0.534 & 0.479 & 8.524 & 0.513 & \textbf{0.791} & 0.152 \\
        & & VACE & 1.039 & 12.783 & 0.545 & 0.593 & 9.572 & 0.519 & 0.755  & 0.178 \\
        & & \cellcolor[rgb]{0.9,0.95,1}Ours & 
        \cellcolor[rgb]{0.9,0.95,1}\textbf{0.530} & 
        \cellcolor[rgb]{0.9,0.95,1}11.475 & 
        \cellcolor[rgb]{0.9,0.95,1}\textbf{0.548} & 
        \cellcolor[rgb]{0.9,0.95,1}\textbf{0.273} & 
        \cellcolor[rgb]{0.9,0.95,1}\textbf{8.349} & 
        \cellcolor[rgb]{0.9,0.95,1}\textbf{0.526} & 
        \cellcolor[rgb]{0.9,0.95,1}0.786 & 
        \cellcolor[rgb]{0.9,0.95,1}\textbf{0.148} \\
    \bottomrule
    \end{tabular}
    \caption{This table compares the performance of different models on full-shot and close-up videos, evaluated at different video resolutions using various metrics. The subscripts `u' and `p' respectively represent the unpaired setting and paired setting. The best results are in bold.}
    \label{Table:compare_common}
\end{table*}

\section{Experiments}

\subsection{Implementation Details}

We conduct all experiments on the Eevee dataset. To facilitate a direct comparison, we performed the experiments at resolutions of $1088\times816$ and $832\times624$. Although the videos in the dataset are all longer than 49 frames, we use the first 49 frames of each video as input and generate 49 output frames. Our experiments are designed in two configurations: paired and unpaired. The paired configuration uses the person's own garment as input to test the model's fidelity in preserving person-specific features. The unpaired configuration, which simulates a realistic use case, challenges the model to dress the person in a different target garment. Besides, we performed virtual try-on on the full-shot videos and the close-up videos, respectively.

For implementation, we initialize our model with the pre-trained weights from VACE and fine-tune it using LoRA~\cite{Lora}. The AdamW~\cite{AdamW} optimizer is used with a fixed learning rate of 1e-4. All training are conducted on 8 NVIDIA H20 (96GB) GPUs. The number of inference steps during testing is set to 50. Furthermore, to ensure a fair comparison, all model variants in the ablation study are trained under identical hyperparameter configurations. 

For evaluation, we employ SSIM~\cite{SSIM} and LPIPS~\cite{LPIPS}. To assess video quality, we use VFID~\cite{VFID} to evaluate both spatial quality and temporal consistency. We report results using two different backbones: I3D~\cite{I3D} ($\text{VFID}_\text{I}$) and ResNeXt~\cite{3D-ResNeXt101} ($\text{VFID}_\text{R}$). In addition, we utilize select metrics from VBench~\cite{Vbench} to comprehensively evaluate video quality and consistency. Finally, to evaluate the consistency between the generated video and the input garment, we report $\text{VGID}$. Due to space limitations, we report the weighted average results across our test categories (dresses, upper body, lower body). Detailed per-category results are in the Appendix.

\subsection{Quantitative Comparison}

\begin{figure}
    \centering
    \begin{subfigure}{0.49\linewidth}
        \includegraphics[width=\linewidth]{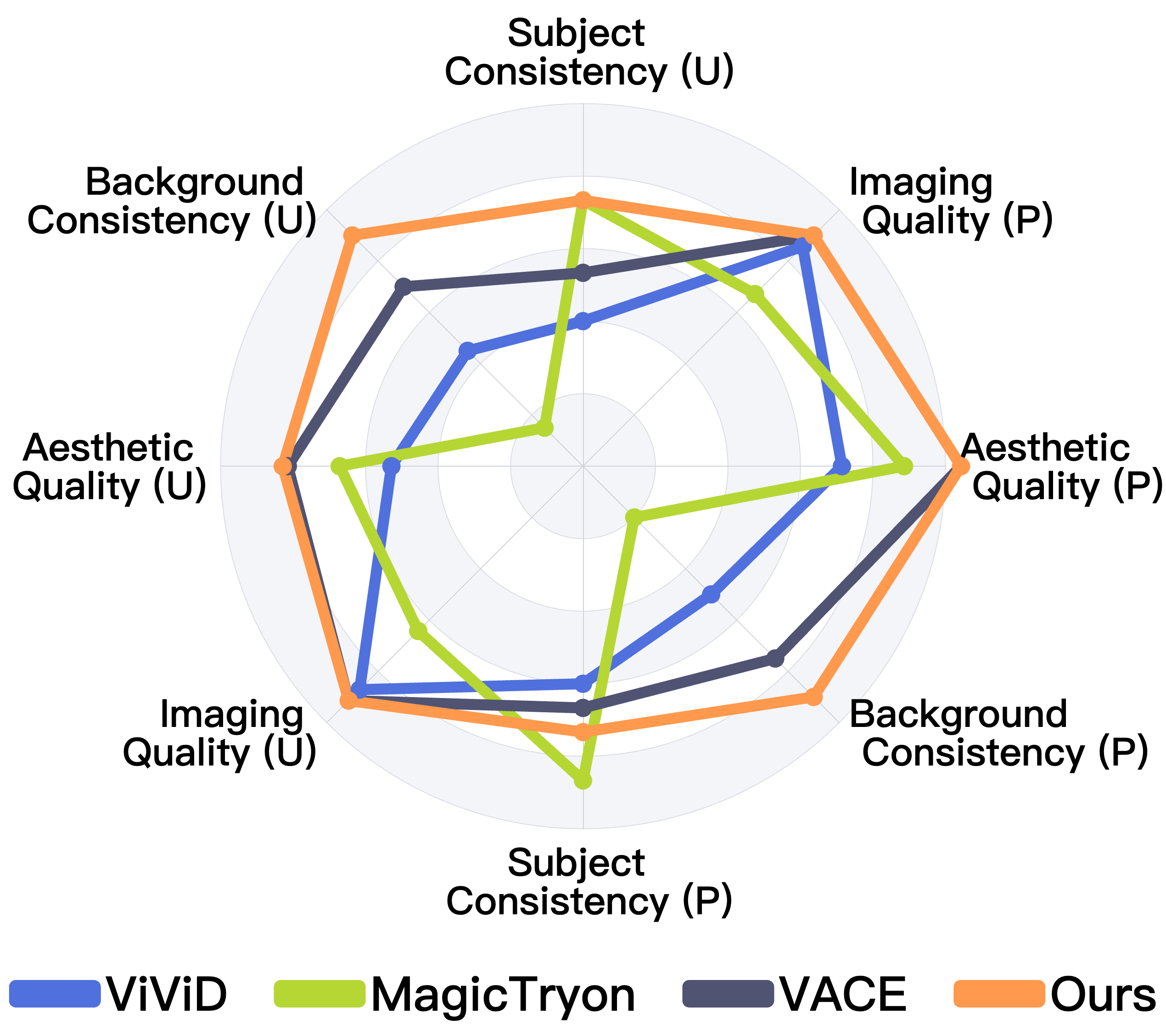}
        \caption{Full-shot video with $1088\times816$ resolution.}
        \label{Fig:vbench_a}
    \end{subfigure}
    \hfill
    \begin{subfigure}{0.49\linewidth}
        \includegraphics[width=\linewidth]{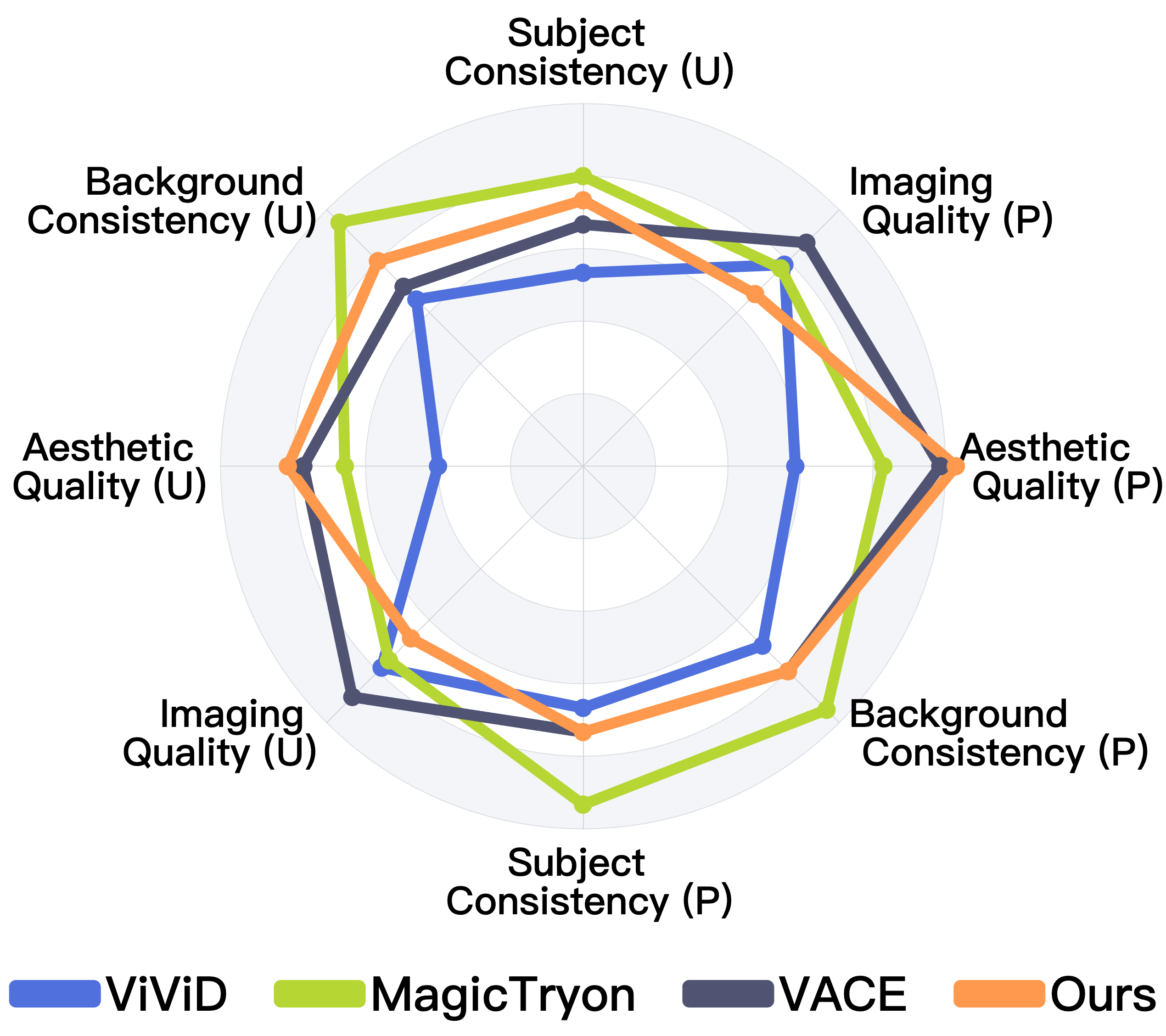}
        \caption{Full-shot video in $832\times624$ resolution.}
        \label{Fig:vbench_b}
    \end{subfigure}
    \hfill
    \begin{subfigure}{0.49\linewidth}
        \includegraphics[width=\linewidth]{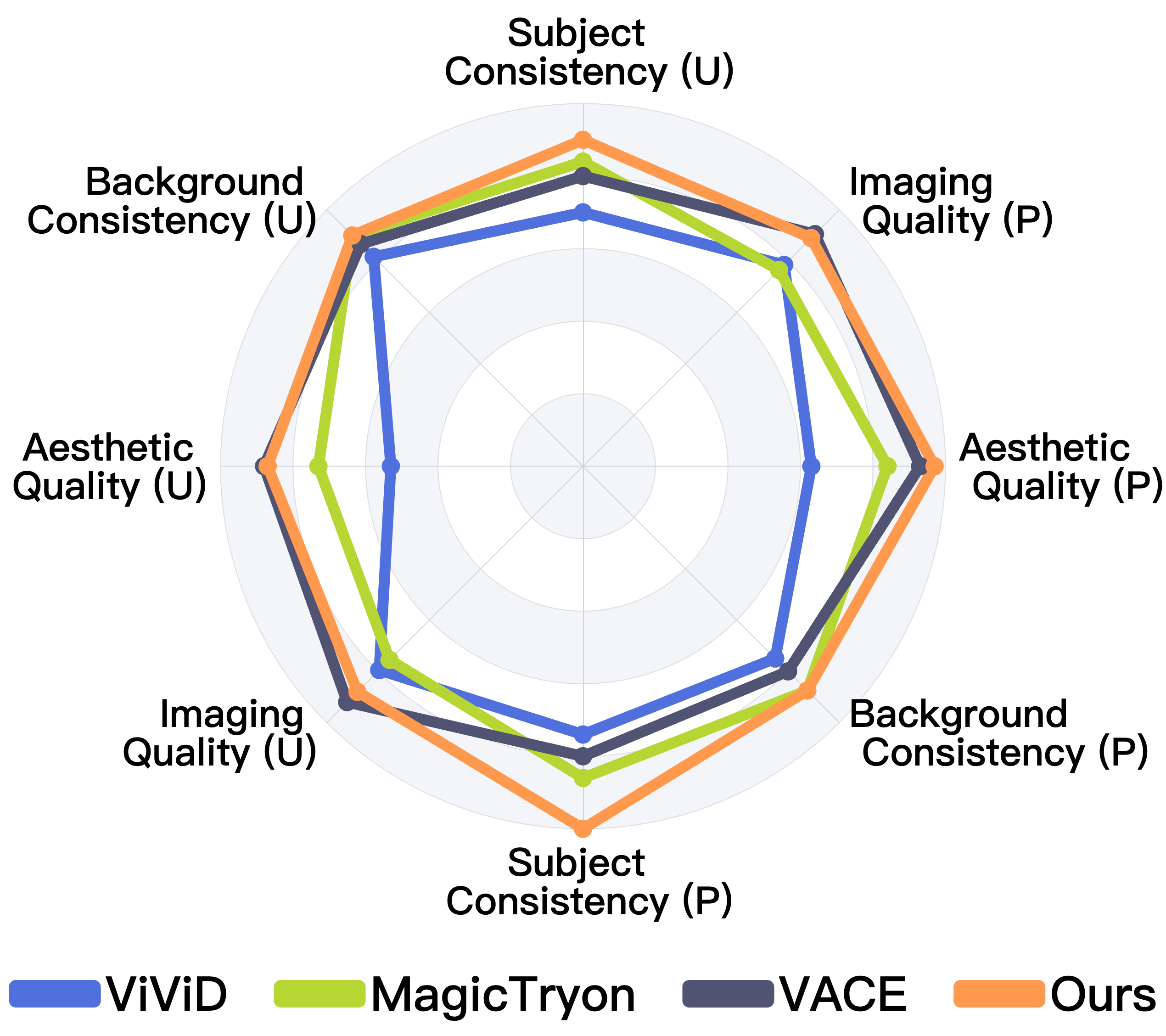}
        \caption{Close-up video in $1088\times816$ resolution.}
        \label{Fig:vbench_c}
    \end{subfigure}
    \hfill
    \begin{subfigure}{0.49\linewidth}
        \includegraphics[width=\linewidth]{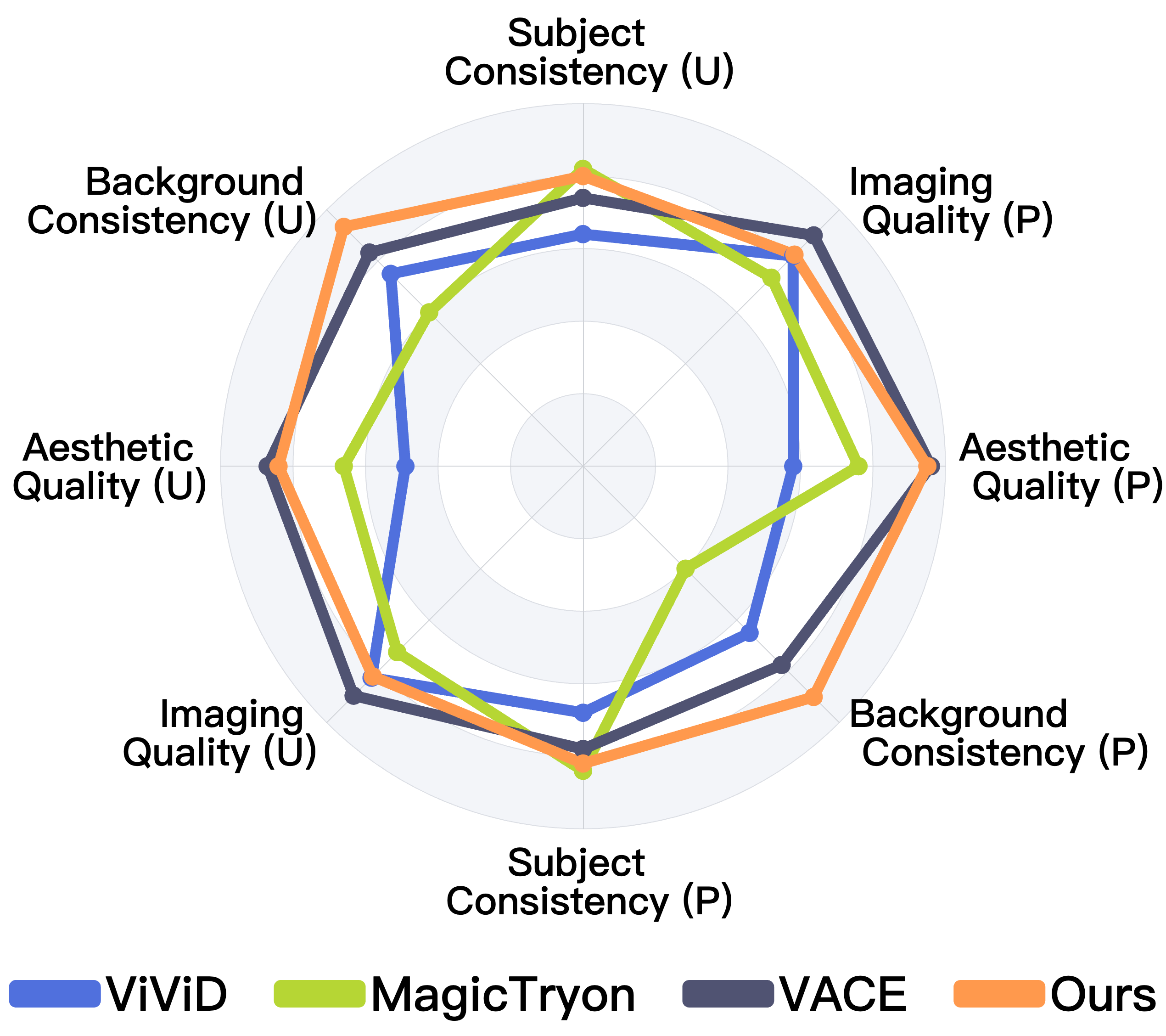}
        \caption{Close-up video in $832\times624$ resolution.}
        \label{Fig:vbench_d}
    \end{subfigure}
    \caption{VBench evaluation results of video-based virtual try-on models. We visualize the evaluation results of four models in 4 VBench dimensions under both paired and unpaired settings. We normalize the results per dimension for clearer comparisons. For comprehensive numerical results, please refer to the Appendix.}
    \label{Fig:vbench}
\end{figure}

\begin{figure*}[t]
    \centering
    \includegraphics[width= 0.96\linewidth]{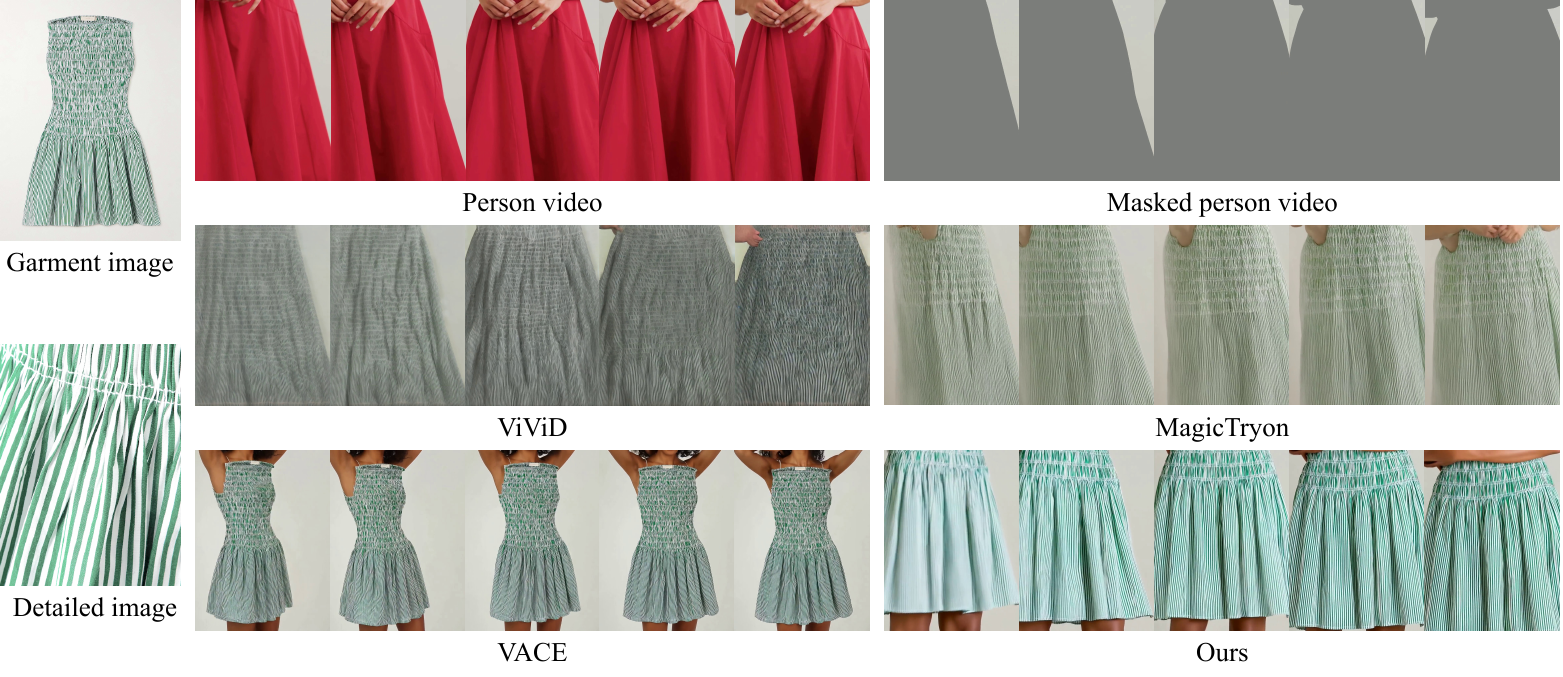}
    \caption{Qualitative comparison of close-up video virtual try-on results on the Eevee dataset. Our fine-tuned VACE model not only accurately reproduces the patterns of the input garment, but the generated video is also more realistic overall. Please zoom in for better visualization.}
    \label{Fig:Qualitative}
\end{figure*}

All of our quantitative comparisons are performed on the Eevee dataset, as our fine-tuned VACE model requires additional garment detail images as input. To ensure a comprehensive and fair evaluation, we designed our experiments across several distinct settings.

As shown in \cref{Table:compare_common}, both MagicTryon~\cite{MagicTryon} and our fine-tuned VACE model~\cite{VACE} achieved the best results. While MagicTryon demonstrated superior performance on the $\text{VFID}_\text{I}$ and LPIPS~\cite{LPIPS} metric, our fine-tuned VACE model excelled on the other metrics, particularly in the paired setting.
We attribute this advantage to the fine-tuning process and detailed garment image. This stage specifically enhanced the model's ability to learn the intricate relationship between the paired inputs, allowing it to better preserve garment details while realistically warping it onto the target pose.
Additionally, the performance of all models decreased to some extent, which is attributable to the increased difficulty of high-resolution and close-up virtual try-on video generation tasks.

Additionally, we utilize the VBench~\cite{Vbench} benchmark to comprehensively evaluate the overall quality and consistency of the generated videos. As shown in \cref{Fig:vbench}, our fine-tuned VACE model secured only a marginal lead. We observed that the metrics across most models showed no significant differentiation. This suggests that the VBench metrics are more heavily influenced by the inherent nature of the task itself rather than the specific capabilities of the models. Consequently, VBench appears limited in its ability to effectively measure or distinguish the true performance differences among various virtual try-on models.

\subsection{Qualitative Comparison}

In \cref{Fig:Qualitative}, we present the visual results of different methods on the Eevee dataset. We observe that our fine-tuned VACE~\cite{VACE} model achieves outstanding performance in generating try-on videos. We attribute this enhanced fidelity directly to the detailed garment images utilized as input during the fine-tuning process, which provide the model with the rich visual information necessary to preserve these specific attributes. Furthermore, this is consistent with the previously mentioned performance on the $\text{Eevee}$. 
ViViD~\cite{VIVID} and MagicTryon~\cite{MagicTryon} often struggle to accurately preserve the patterns of the input garment. The original VACE tends to directly paste the garment onto the target area rather than performing a genuine try-on. As shown in \cref{Fig:Qualitative}, the tag on the garment's collar in the VACE result should be hidden behind the model.
Additionally, since VACE is not optimized for video-based virtual try-on, it occasionally experiences generation failures, such as producing solid white clothing or failing to generate human body parts entirely. This leads to poor quantitative metrics for VACE.

\begin{figure*}[t]
  \centering
  \includegraphics[width=\linewidth]{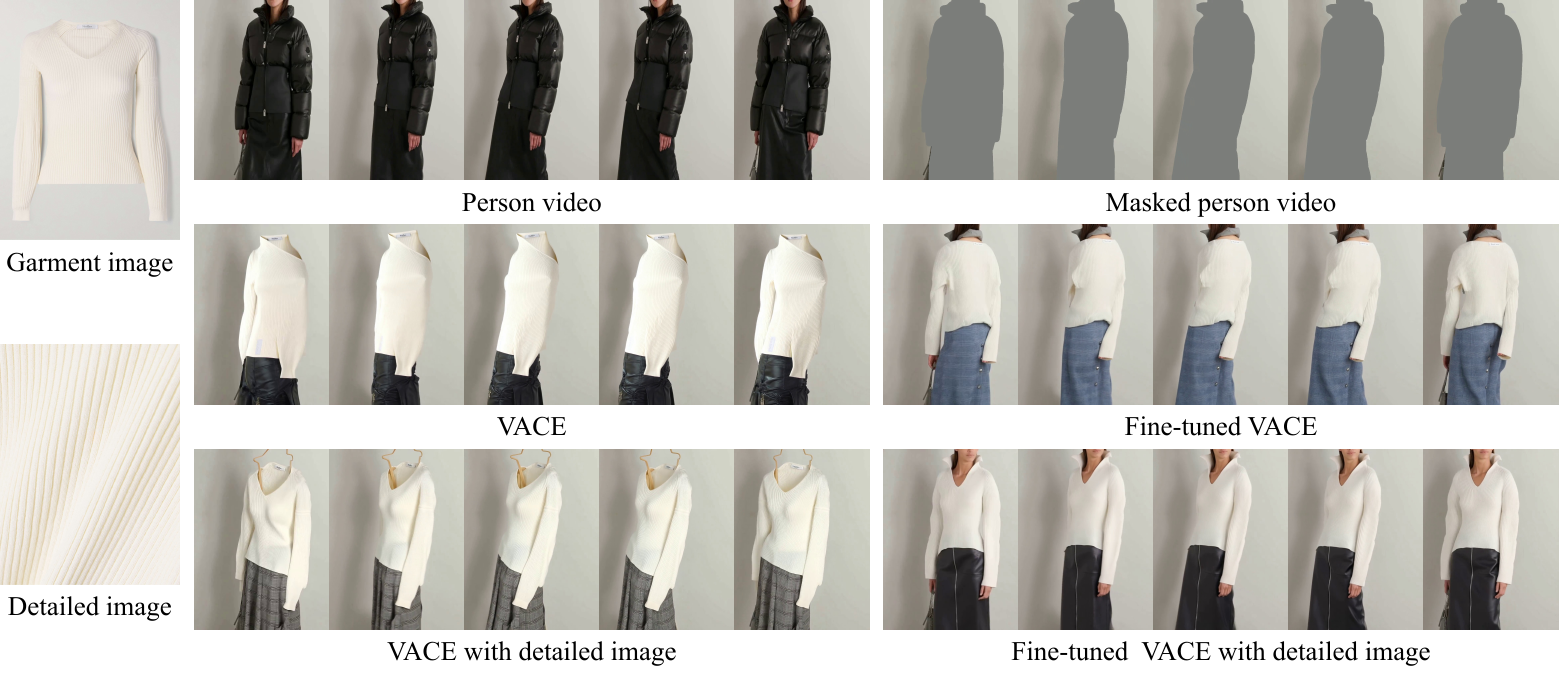}
   \caption{Qualitative comparison of the ablation study. The model combining LoRA fine-tuning and detailed images generates textures with significantly enhanced clarity and fidelity compared to other models. Please zoom in for better visualization}
   \label{Fig:Ablation}
   \vspace{3mm}
\end{figure*}

\begin{table*}[t]
    \footnotesize
    \centering
    \setlength{\tabcolsep}{5.1pt}
    
    \begin{tabular}{ll cc ccc ccccc}
    \toprule
        \multirow{2}{*}[-3px]{\textbf{Resolution}} & 
        \multirow{2}{*}[-3px]{\textbf{Type}} & 
        \multirow{2}{*}[-3px]{\textbf{\makecell{LoRA\\Finetune}}} & 
        \multirow{2}{*}[-3px]{\textbf{\makecell{Detailed\\Image}}} & 
        \multicolumn{3}{c}{\textbf{Unpaired}} &
        \multicolumn{5}{c}{\textbf{Paired}} \\
        \cmidrule(lr){5-7} 
        \cmidrule(lr){8-12} 
         & & & & \textbf{$\text{VFID}_\text{R}^\text{u}\downarrow$ } & \textbf{$\text{VFID}_\text{I}^\text{u}\downarrow$ } & 
         \textbf{$\text{VGID}^\text{u}\uparrow$ } &
         \textbf{$\text{VFID}_\text{R}^\text{p}\downarrow$ } & \textbf{$\text{VFID}_\text{I}^\text{p}\downarrow$ } & 
         \textbf{$\text{VGID}^\text{p}\uparrow$ } &
         \textbf{SSIM$\uparrow$} &
         \textbf{LPIPS$\downarrow$} \\
    \midrule
        \multirow{4}{*}{$1088\times816$} & \multirow{4}{*}{Full-shot} & \ding{55} & \ding{55} & 0.454 & 11.260 & 0.522 & 0.282 & 7.290 & 0.525 & 0.899 & 0.083 \\ 
        & & \ding{55} & \ding{51} & 0.661 & 11.290  & 0.523 & 0.400 & 7.380  & 0.531 & 0.896 & 0.084 \\
        & & \ding{51} & \ding{55} & 0.225 & 10.219 & 0.520 & 0.093 & 6.558  & 0.529 & 0.909 & 0.079 \\
        & & \cellcolor[rgb]{0.9,0.95,1}\ding{51} & \cellcolor[rgb]{0.9,0.95,1}\ding{51} & 
        \cellcolor[rgb]{0.9,0.95,1}\textbf{0.220} &
        \cellcolor[rgb]{0.9,0.95,1}\textbf{10.065} & 
        \cellcolor[rgb]{0.9,0.95,1}\textbf{0.524} & 
        \cellcolor[rgb]{0.9,0.95,1}\textbf{0.076} & 
        \cellcolor[rgb]{0.9,0.95,1}\textbf{6.555} & 
        \cellcolor[rgb]{0.9,0.95,1}\textbf{0.534} & 
        \cellcolor[rgb]{0.9,0.95,1}\textbf{0.910} & 
        \cellcolor[rgb]{0.9,0.95,1}\textbf{0.078} \\
    \midrule
        \multirow{4}{*}{$1088\times816$} & \multirow{4}{*}{Close-up} & \ding{55} & \ding{55} & 0.913 & 12.844 & 0.533 & 0.429 & 9.331 & 0.515 & 0.778 & 0.174  \\
        & & \ding{55} & \ding{51} & 1.218 & 12.666 & 0.525 & 0.628 & 9.162  & 0.520 & 0.770 & 0.174 \\
        & & \ding{51} & \ding{55} & 0.608 & 11.757 & 0.529 & \textbf{0.220} & 8.441  & 0.517 & 0.802 & 0.156 \\
        & & \cellcolor[rgb]{0.9,0.95,1}\ding{51} & \cellcolor[rgb]{0.9,0.95,1}\ding{51} & 
        \cellcolor[rgb]{0.9,0.95,1}\textbf{0.535} & 
        \cellcolor[rgb]{0.9,0.95,1}\textbf{11.621} & 
        \cellcolor[rgb]{0.9,0.95,1}\textbf{0.540} & 
        \cellcolor[rgb]{0.9,0.95,1}0.274 & 
        \cellcolor[rgb]{0.9,0.95,1}\textbf{8.429} & 
        \cellcolor[rgb]{0.9,0.95,1}\textbf{0.524} & 
        \cellcolor[rgb]{0.9,0.95,1}\textbf{0.803} & 
        \cellcolor[rgb]{0.9,0.95,1}\textbf{0.154} \\
        
    \bottomrule
    \end{tabular}
    \caption{Ablation study of LoRA fine-tuning and detailed images across full-shot and close-up videos. The subscripts `u' and `p' respectively represent the unpaired setting and paired setting. The best results are in bold.}
    \label{Table:ablation}

\end{table*}

\subsection{Ablation Study}

To better understand the impact of detailed garment images and the fine-tuning process, we conducted a series of comparative experiments based on the VACE model. We compared four VACE~\cite{VACE} model configurations: the original model, the model using detailed garment images as additional input, the model fine-tuned on the Eevee dataset with LoRA~\cite{Lora}, and the model both fine-tuned with LoRA and using the additional detailed garment images.

As shown in ~\cref{Table:ablation}, our experimental data reveals that the impact of directly inputting detailed garment images alone is negligible, with the model showing performance comparable to the original VACE baseline. This suggests the model struggles to leverage this extra information without specific adaptation. Conversely, the value of these detailed images becomes apparent during fine-tuning. Specifically, the model fine-tuned with LoRA while using the additional detailed garment images achieves superior results compared to the model fine-tuned with LoRA without them. This strongly indicates that the fine-tuning process is crucial for enabling the model to effectively exploit the high-fidelity information present in the detailed images.

This conclusion is further corroborated by the qualitative visual comparisons presented in \cref{Fig:Ablation}. The original VACE model produced incorrect results, generating videos that featured only the garment with no person. While providing additional detailed images of the garment made the output more natural, the human body was still missing.
In contrast, the VACE model, after being fine-tuned on the Eevee dataset, successfully generated a human. Furthermore, the fine-tuned VACE model, when combined with the detailed garment image, not only produced a human body but also achieved a more natural-looking result overall.

Finally, recognizing that distribution metrics can sometimes be inconsistent with human perception, we conducted a user study (N=53) focusing on semantic alignment. Raters preferred our configuration with detailed guidance over the baseline without it in preserving texture (\textbf{79.25\%}), pattern (\textbf{83.02\%}), and overall quality (\textbf{84.91\%}). This subjective evaluation confirms that the detailed image is critical for accurately preserving specific garment identity.

\section{Conclusion}


In this paper, we addressed the critical limitations of existing video-based virtual try-on: low resolution, only full-shot videos, and the failure to capture fine-grained garment textures from single input garment images. Our primary contribution is the introduction of Eevee, a high-resolution dataset designed to support a wide range of virtual try-on tasks. It is the first dataset to provide both full-shot and close-up videos, and corresponding detailed close-up images for the garments. Recognizing that this task requires better evaluation, we also proposed a new metric VGID to specifically measure garment consistency. Leveraging this dataset and metric, we conducted a comprehensive benchmark of recent state-of-the-art models, establishing a baseline for future research in virtual try-on. We hope our work inspires further research into virtual try-on.

{
    \small
    \bibliographystyle{ieeenat_fullname}
    \bibliography{main}
}


\end{document}